# Beyond Gröbner Bases: Basis Selection for Minimal Solvers

Viktor Larsson
Lund University
Lund, Sweden
viktorl@maths.lth.se

Magnus Oskarsson
Lund University
Lund, Sweden
magnuso@maths.lth.se

Kalle Åström
Lund University
Lund, Sweden
kalle@maths.lth.se

Alge Wallis
University of Bath
Bath, United Kingdom
d.wallis@bath.ac.uk

Zuzana Kukelova
FEE
kukelzuz@fel.cvut.cz

Tomas Pajdla
CIIRC
pajdla@cvut.cz

Czech Technical University in Prague, Czech Republic

## Abstract

*Many computer vision applications require robust estimation of the underlying geometry, in terms of camera motion and 3D structure of the scene. These robust methods often rely on running minimal solvers in a RANSAC framework. In this paper we show how we can make polynomial solvers based on the action matrix method faster, by careful selection of the monomial bases. These monomial bases have traditionally been based on a Gröbner basis for the polynomial ideal. Here we describe how we can enumerate all such bases in an efficient way. We also show that going beyond Gröbner bases leads to more efficient solvers in many cases. We present a novel basis sampling scheme that we evaluate on a number of problems.*

## 1. Introduction

In this paper, we describe a method for automatically building very fast minimal solvers. This is a core problem in many computer vision applications, *e.g.* 3D reconstruction [41, 42], visual odometry [35, 2] and visual localization [37, 47].

During the last years we have seen a large increase in the variety of available technical platforms, such as mobile devices, UAVs and drones. These are often equipped with widely different capabilities, in terms of sensors, cameras and computing power. In many applications, *e.g.* autonomous navigation, augmented reality or robotics, a core computer vision task is to make robust estimates of the surrounding geometry and motion of the device, based on image data [50, 11, 47, 48, 38] or other sensor data [23, 40, 3, 22]. Given that these tasks often need to be performed fast, based on unreliable data containing mismatches, and on devices with limited processing power, efficient implementations of robust estimation schemes such as RANSAC is paramount. At the core of these algorithms we have so-called minimal solvers that, based on a small data sample, estimate a model that can be evaluated on a larger data set to find a consistent inlier set. Since this has to be performed many times [14], we need these minimal solvers to be fast. The image formation process naturally leads to geometric problems that can be formulated in terms of multivariate polynomial equation systems. In many devices we have additional sensor measurements, *e.g.* gyroscope and accelerometer data, and these measurements should be incorporated in the estimation process [1, 18, 47]. We can also have different types of camera models and calibration knowledge [50, 10, 49, 33, 30, 48, 17, 38]. All the aspects described above, naturally lead to the need for tools for constructing fast solvers of polynomial equations. In addition, the variety of platforms leads to the need for these methods to be, to a large extent, automatic.

Many state-of-the-art polynomial solvers are based on Gröbner bases and the action-matrix method, and there are now powerful tools available for the automatic generation of such solvers [25, 28, 29]. In this paper we target a specific part of this pipeline, namely the choice of monomial basis in the quotient ring. We will show how careful selection of the monomial bases can give significant speed-up in the resulting solvers. Previously, little attention has been paid to the choice of basis to gain speed in polynomial solvers, and usually a Gröbner basis is used to select the monomial basis. We will in the paper describe how we can test *all possible* Gröbner bases. We will further show that going beyond Gröbner bases leads to faster solvers in a number of cases.

Specifically, our contributions in this paper are:

- Minimizing elimination template size by enumerating all possible Gröbner bases.

- A heuristic method of sampling monomial bases that goes beyond Gröbner bases.
- State-of-the-art performance on a number of geometric estimation and calibration problems in terms of speed.

### 1.1. Background and Related Work

We use nomenclature and basic concepts from Cox *et al.* [12]. Let $X = (x_1, x_2, \ldots, x_n)$ denote our indeterminates and let $\mathbb{C}[X]$ denote the set of all polynomials in $X$ with coefficients in $\mathbb{C}$. We are interested in solving systems of polynomial equations,

$$\begin{cases} f_1(x_1, \ldots, x_n) = 0, \\ \vdots \\ f_m(x_1, \ldots, x_n) = 0. \end{cases} \quad (1)$$

The ideal $I = \langle f_1, \ldots, f_m \rangle$ is the set of polynomial combinations of our generators $f_1, \ldots, f_m$. Each polynomial $f \in I$ then vanishes on the solutions to our equation system (1). Using the ideal $I$ we can define the quotient ring $\mathbb{C}[X]/I$ which is the set of equivalence classes over $\mathbb{C}[X]$ defined by the relation,

$$a \sim b \iff a = b \mod I \iff a - b \in I. \quad (2)$$

If there are finitely many solutions to (1), then the quotient ring $\mathbb{C}[X]/I$ is a finite-dimensional vector space over $\mathbb{C}$.

For an ideal $I$ there exist special sets of generators called Gröbner bases which have the nice property that the remainder after division is unique. For a Gröbner basis $\{g_1, \ldots, g_m\}$ we can define the standard monomials, which is the set of monomials not divisible by the leading term of any $g_k$. This set of monomials is a linear basis for the quotient ring $\mathbb{C}[X]/I$.

### 1.2. Solving Systems of Polynomial Equations

To solve systems of polynomial equations, the most common approach in Computer Vision is the action matrix method [13, 46, 43, 27]. The goal of the action matrix method is to transform the, in general very difficult, problem of finding the solutions to an equivalent eigenvalue/eigenvector problem which we can solve numerically:

$$\alpha \begin{bmatrix} b_1 \\ \vdots \\ b_K \end{bmatrix} - \begin{bmatrix} & & \\ & M & \\ & & \end{bmatrix} \begin{bmatrix} b_1 \\ \vdots \\ b_K \end{bmatrix} = 0 \mod I, \quad (3)$$

where $\alpha, b_k \in \mathbb{C}[X]$. Here $\alpha$ is the so-called *action variable*. To briefly motivate this, consider the quotient ring $\mathbb{C}[X]/I$ of an ideal $I$ generated by polynomials with a finite number of solutions $K$. Let $b_1, \ldots, b_K \in \mathbb{C}[X]$ be monomials forming a basis of $\mathbb{C}[X]/I$. Then, the remainder of the product $\alpha b_i$ after division by $I$ can be written as a linear combination of the basis $b_1, \ldots, b_K \in \mathbb{C}[X]$ [13]. In matrix form we then get (3). The existence of the matrix $M \in \mathbb{C}^{K \times K}$ can be guaranteed by choosing $b_1, \ldots, b_K$ as a basis for the quotient ring $\mathbb{C}[X]/I$.

In practice, to recover the action matrix $M$ for a particular instance a so-called *elimination template* is typically used. The elimination template is an expanded set of equations (constructed by multiplying the original equations with different monomials) in which we can linearly express the polynomials in the eigenvalue problem. This reduces the problem of finding the action matrix to solving a linear system.

It is generally difficult to find the smallest elimination template for a given problem. In [25] an iterative method was presented for constructing these elimination templates directly from the problem equations. This work was recently extended by Larsson *et al.* in [28], where a non-iterative method was proposed.

When creating polynomial solvers in Computer Vision, the basis for the quotient space $\{b_1, \ldots, b_K\}$ is typically chosen as the standard monomials from the Gröbner basis w.r.t. the monomial ordering GRevLex (this is e.g. done in [25, 28]). However, this is an arbitrary choice and the methods work for any basis. In this paper we focus on the problem of selecting this basis with the aim of reducing the size of the elimination template. We show that for some problems there are better choices that yield significantly faster solvers.

Given a monomial basis it is still a difficult problem to find the smallest elimination template. In this work we use the automatic generator from [28] to construct the templates, but this method is not guaranteed to find the optimal template. The results in this paper are w.r.t. this particular template construction method. However, any other method for constructing the template (such as the one from [25]) could be substituted.

Our approach can be used to optimize other characteristics of the solvers as well, such as accuracy or stability. However, for practical purposes these are typically secondary to runtime as long as they are sufficiently good. For example if you are estimating the focal length it usually does not matter whether the errors are $10^{-6}$ or $10^{-16}$.

In [9] the authors presented different methods for choosing the basis during runtime. However in their setting the size of the template was fixed, and the online basis selection was done solely to improve the numerics of the solver.

## 2. Exhaustive Search over Gröbner Bases

For an ideal $I$, the (reduced) Gröbner basis depends on the monomial ordering chosen in the polynomial ring $\mathbb{C}[X]$. Different orderings can yield different Gröbner bases, and thus different sets of standard monomials. For polynomial solvers in Computer Vision, the most popular ordering is GRevLex [12], since it has been empirically observed to typically give small elimination templates [44, 43, 25, 28]. It has also been noted in computational algebraic geometry and cryptography [39] that graded orderings [12] (*i.e.* archimedean [36]) often lead to faster Gröbner basis computations compared to, *e.g.* lexicographical orderings [12]. However, there exist examples where this is not the case. Hence, this suggests to investigate the efficiency of Gröbner basis (and hence action matrix) construction w.r.t. all possible different monomial orderings.

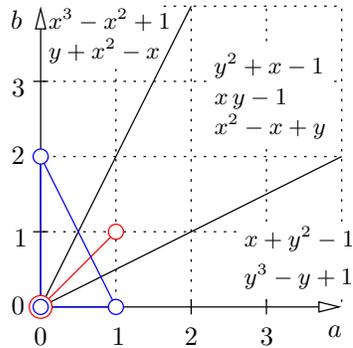

Figure 1. The Gröbner fan of the ideal $I = \langle x + y^2 - 1, x\,y - 1\rangle$ consists of three two-dimensional cones. For each cone, there is exactly one reduced Gröbner basis of $I$. All monomial orderings generated by all weight vectors from one cone give the same reduced Gröbner basis of $I$. Hence, there are exactly three different reduced Gröbner bases for $I$ over all possible different monomial orderings.

### 2.1. Gröbner Fans

While there are (uncountably) infinitely many different monomial orderings, Mora and Robbiano [32] showed that for a given ideal $I$ there are only finitely many different reduced Gröbner bases [16]. To present this theory is beyond the scope of this paper, but we will try to describe the main ideas and relate how this can be used in our problem setting. The set of all reduced Gröbner bases of an ideal can be computed [16, 20, 19] using the *Gröbner fan* of the ideal [32, 45]. The Gröbner fan of an ideal was defined by Mora and Robbiano in 1988 [32]. It is a finite fan of polyhedral cones indexing the distinct monomial initial ideals with respect to monomial orderings or, equivalently, indexing the reduced Gröbner bases of the ideal. See [32, 45, 16] for the full account of the theory.

Here we will illustrate it on a simple example computed using the software package Gfan [19, 20]. Consider the polynomial system $I = \langle x + y^2 - 1, x\,y - 1\rangle$. Figure 1 shows the Gröbner fan of $I$ together with the corresponding reduced Gröbner bases. It consists of three two-dimensional cones. For each cone, there is exactly one reduced Gröbner basis of $I$, giving in total three different reduced Gröbner bases. To connect the different reduced Gröbner bases to the fans in Figure 1, consider the exponent vectors $[a, b]^\top$ that correspond to monomials $x^a y^b$, *e.g.* $[2, 3]^\top$ represents $x^2 y^3$. Now, for every monomial ordering $\prec$ on $\mathbb{C}[x, y]$ one can find a (set of) real non-negative (weight) vectors $w \in \mathbb{R}^2$ such that if $x^a y^b \prec x^c y^d$, then $[a, b] \cdot w \leq [c, d] \cdot w$. In this way, every ordering is connected to a set of its (compatible) real weight vectors. Finally, for a fixed $I$, the union of all the sets of weight vectors corresponding to all monomial orderings producing the same reduced Gröbner basis is a full (here two) dimensional cone in $\mathbb{R}^2$. There are only finitely many such cones for a fixed $I$. In our situation, there are three two-dimensional cones, see Figure 1.

### 2.2. Building Minimal Solvers using Gröbner Fans

In [28], the state-of-the-art automatic generator for polynomial solvers has been presented and evaluated on a large test-bed of polynomial equation systems from geometric Computer Vision. Even though some of the problem formulations from [28] are no longer state-of-the-art for their respective problem, they still serve as a good benchmark set to test our methods. For each of these problems we tried to compute the Gröbner fan, aborting the computations if they lasted more than 12 hours. Using the automatic generator we then construct a polynomial solver for each of the reduced Gröbner bases found. Table 1 shows some problems where we were able to find a smaller elimination template compared to using the GRevLex basis. Note that the number of Gröbner bases can increase very quickly and it is not always tractable to compute the complete Gröbner fan for larger problems. For the six point relative pose with shared radial distortion problem we ran the Gröbner fan computation for a week before aborting the computation.

Figure 2 shows a histogram of the different template sizes for the solvers constructed from the Gröbner fan for the P4Pfr formulation from Bujnak *et al.* [7]. We can see that many of the found bases give very large templates. To avoid these uninteresting bases, as well as the long runtimes for computing the Gröbner fan, we propose to use a guided random sampling approach in the next section.

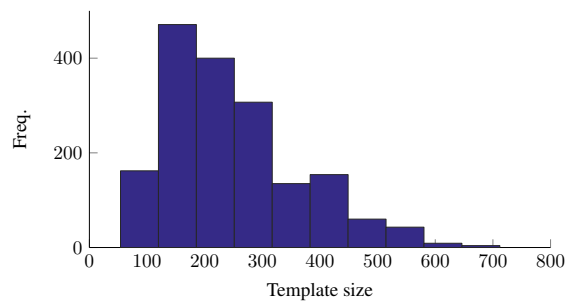

Figure 2. Template size (rows) for the Gröbner fan bases for the P4Pfr formulation from Bujnak *et al.* [7].

## 3. Beyond Gröbner Bases

In the previous section we computed all reduced Gröbner bases for a problem and used these to select quotient ring bases. However, it is not necessary to select a standard monomial basis that comes from a Gröbner basis for some monomial ordering, since any spanning and linearly independent set will do. In this section we instead consider bases which do not come as standard monomial bases from any Gröbner basis, and show that for some problems we can find even smaller elimination templates.

### 3.1. Random Sampling for Basis Selection

Once you drop the Gröbner basis constraint you have infinitely many choices for monomial bases, so it is no longer possible to do any exhaustive search. Even if we restrict ourselves to monomials below some fixed degree, the combinatorial explosion of choices often makes it intractable to try them all.

Instead we propose a random sampling approach. The sampling is guided by several heuristics based on empirical observations. We will motivate our choices later on, but we will start by describing our proposed algorithm. The heuristics we use for basis selection are:

H1. We try to have as many of the basis monomials and reducible monomials (*i.e.* $\alpha b_k$) appearing in the original equations as possible.

H2. We try to minimize the degree in some of the unknowns. This usually helps when the variables occur in an unbalanced way in the equations. E.g. if our problem is parameterized using a quaternion (for rotation) and a focal length, we have seen that it is typically good to try to minimize the degree of the focal length.

H3. We try to select a connected block of monomials.

To generate the initial set of monomials that we sample from we use the following strategy: We start with the monomials occurring in the equations. If these do not contain any basis (see Section 3.2) we multiply with all first degree monomials that occur in the equations and add these. If they still do not contain any basis, we again multiply with all the second degree monomials and so on (in some special cases we need to add some extra low-degree monomials to get an independent set). We denote these monomials by $\mathcal{M}$ and the monomials that occur in the original equations by $\mathcal{E} \subset \mathcal{M}$.

Now, to sample a basis we start by randomly choosing a binary weight vector $\omega = \{0,1\}^n$. This represents the direction we want to minimize in H2. For each monomial $m \in \mathcal{M}$ we assign a weight $w_d(m)$ penalizing the weighted degree using $\omega$. So, e.g. if $\omega = (0,1,1)$, the monomial $m = xyz^2$ would have the weighted degree $0 + 1 + 2 = 3$. Next we select an action variable $\alpha$. It is chosen uniformly in the direction which is minimized by $\omega$. So, in the previous example we would have chosen either $y$ or $z$. If $\omega$ is all zero we choose uniformly from all variables. Note that this $\alpha$ is used only for guiding the random sampling. When we construct the solvers we try every variable as action.

The basis is then sampled iteratively, with one monomial added at a time. Given a partial basis $\mathcal{B} \subset \mathcal{M}$ we select the next monomial to add as follows:

1. Find monomials $\mathcal{M}_\mathcal{B} \subset \mathcal{M}$ that are linearly independent from the partial basis $\mathcal{B}$ (see Section 3.2)

2. For each monomial $m \in \mathcal{M}_\mathcal{B}$ compute a weight
$$w(m) = \mathbb{I}(m \in \mathcal{E}) + \mathbb{I}(\alpha m \in \mathcal{E} \cup \mathcal{B}) + w_d(m) + \epsilon \quad (4)$$
where $\epsilon$ is a small number.

3. Find the neighboring monomials of $\mathcal{B}$ in $\mathcal{M}_\mathcal{B}$.

4. Sample proportionally to $w(m)$ from the neighboring monomials. (If there are no neighboring monomials in $\mathcal{M}_\mathcal{B}$, sample instead from all of $\mathcal{M}_\mathcal{B}$).

These steps are iterated until we have a complete basis.

### 3.2. Checking Linear Independence

When we sample basis elements, we need to be able to quickly determine if a set of monomials are linearly independent in the quotient ring $\mathbb{C}[X]/I$ (or typically $\mathbb{Z}_p[X]/I$ since we do most of our calculations in $\mathbb{Z}_p$ to speed up computations and avoid round-off errors).

We start by computing any (reduced) Gröbner basis for the ideal and find the standard monomials $\{b_1, b_2, \ldots, b_K\}$ for this basis. Then, since these monomials form a basis for the quotient ring, we write each $m \in \mathcal{M}$ as
$$m = \sum_k c_k b_k \mod I, \quad (5)$$
by simply dividing with the Gröbner basis. This associates vector $\mathbf{c} = (c_1, c_2, \ldots, c_K)$ to each monomial in $\mathcal{M}$. To check if a set of monomials is linearly independent in the quotient ring, we can now equivalently check if the corresponding vectors are independent in $\mathbb{C}^K$ (or $\mathbb{Z}_p^K$) by performing the standard Gaussian elimination.

### 3.3. Building Minimal Solvers with Sampled Bases

We applied our random sampling strategy in an experiment similar to the one in Section 2.2. For each problem, we randomly sampled 100 bases and constructed the corresponding solvers. Some results are shown in Table 1. We can see that using our sampling strategy we can find smaller elimination templates for some problems. Note that for some problems the best basis did not come from any

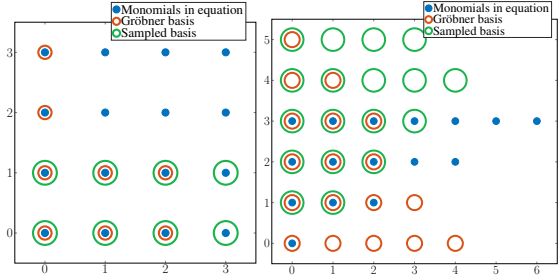

Figure 3. The figure shows the basis monomials for two example problems, namely 8pt rel. pose F+$\lambda$ (left) and 3pt image stitching f$\lambda$+R+f$\lambda$ (right). Both these problems have two variables, and for both these problems the proposed basis sampling scheme gives significantly smaller template compared to the Gröbner basis variants.

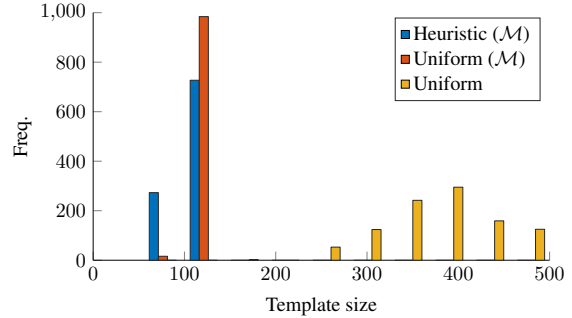

Figure 4. Template size (rows) for 1,000 randomly sampled bases for the P4Pfr formulation from Bujnak *et al.* [7].

Gröbner basis. We were also able to find smaller solvers for problems where the Gröbner fan computation took too long and was aborted.

In Figure 3 we show the best resulting monomial bases from our sampling scheme, for two cases. The monomials are here represented by their corresponding exponents as vectors. We also show the standard GRevLex bases, that give significantly larger templates for these two problems. In general the GRevLex ordering will lead to a basis that has a low total degree. We have found that, for some problems, if it is possible to keep the maximum degree low in one variable, even if the total degree becomes larger, this is beneficial for the template size. Figure 3 left shows an example of this. Another important aspect that we have seen, is that we should choose, if possible, monomials within the original equations, as these are available directly. In Figure 3 right, all the monomials occurring in the original equations are shown as blue dots. In this case the sampled basis better aligns with the structure of the monomials in the equations compared to GRevLex.

### 3.4. Experiment: Heuristic vs. Uniform Sampling

In this section we show a comparison of our heuristic with sampling basis monomials uniformly. We compare three different approaches: (i) our heuristic sampling from the monomials in $\mathcal{M}$ (as defined in Section 3.1), (ii) uniformly sampling from $\mathcal{M}$, and (iii) uniformly sampling from all monomials of the same degree as those in $\mathcal{M}$. Figure 4 shows the distribution of the template sizes (number of rows) for 1,000 random samples for the P4Pfr formulation from Bujnak *et al.* [7]. We can see that our sampling heuristic and the strategy for selecting $\mathcal{M}$ both give significant improvements.

## 4. Panoramic Stitching $f\lambda + R + f\lambda$

We will now show how our method can be used to construct fast solvers for stitching images from cameras with radial distortion and where the focal length is unknown. This problem was formulated and solved using Gröbner-basis techniques in [8]. In [34] a technique for numerically optimizing the size of the elimination template was presented, and a new faster solver with a template of size $54 \times 77$ was constructed. In [28] a slightly faster solver was presented, based on a template of size $48 \times 66$. We will follow the derivations in [5] and [8] when we construct our solver for two-view stitching using three point correspondences. We will additionally show that we can use the exact same solver to solve the minimal problem of three-view stitching using two point correspondences.

### 4.1. Two View Image Stitching

We assume that we have a camera undergoing some unknown rotation $R$, taking two images of a number of unknown 3D points $\boldsymbol{X}_i$. We denote the points in the two images with $\boldsymbol{u}_i$ and $\boldsymbol{u}'_i$ respectively. We will describe how we handle the radial distortion later, and will assume that we only need to handle the unknown focal length $f$ just now. The projection equations can then be formulated as

$$\gamma_i \boldsymbol{u}_i = K\boldsymbol{X}_i, \quad \gamma'_i \boldsymbol{u}'_i = KR\boldsymbol{X}_i, \qquad (6)$$

where $\gamma_i$ and $\gamma'_i$ are the depths, and $K = \mathrm{diag}(f, f, 1)$. We can remove the dependence of $\gamma_i$, $\gamma'_i$ and $R$ by solving for $\boldsymbol{X}_i$ and taking scalar products, giving the constraints

$$\frac{\langle K^{-1}\boldsymbol{u}_j, K^{-1}\boldsymbol{u}_k \rangle^2}{|K^{-1}\boldsymbol{u}_j|^2 |K^{-1}\boldsymbol{u}_k|^2} = \frac{\langle \boldsymbol{X}_j, \boldsymbol{X}_k \rangle^2}{|\boldsymbol{X}_j|^2 |\boldsymbol{X}_k|^2} = \frac{\langle K^{-1}\boldsymbol{u}'_j, K^{-1}\boldsymbol{u}'_k \rangle^2}{|K^{-1}\boldsymbol{u}'_j|^2 |K^{-1}\boldsymbol{u}'_k|^2}, \qquad (7)$$

for two points $j$ and $k$. Cross-multiplying with denominators will give polynomials in the unknown $f$. We will now add radial distortion to our problem, and model it using Fitzgibbon's division model [15] so that for the radially distorted image coordinates $\boldsymbol{x}_i$ we have $\boldsymbol{u}_i \sim \boldsymbol{x}_i + \lambda \boldsymbol{z}_i$, where $\boldsymbol{z}_i = \begin{bmatrix} 0 & 0 & x_i^2 + y_i^2 \end{bmatrix}^T$, and $\lambda$ is the radial distortion parameter. Inserting this into (7) gives us our final constraints in the unknown $\lambda$ and $f$. Using only two points will only give us one equation so we need at least three point correspondences (this actually gives three constraints, so it is slightly over-determined, but we only use two of the equations).

| Problem | Author | Original | [28] | GFan+ [28] | (#GB) | Heuristic+[28] |
|---|---|---|---|---|---|---|
| Rel. pose F+$\lambda$ 8pt | Kuang *et al*. [24] | $12 \times 24$ | $11 \times 20$ | $11 \times 20$ | (10) | $\mathbf{7 \times 16}$ |
| Rel. pose E+$f$ 6pt | Bujnak *et al*. [6] | $21 \times 30$ | $21 \times 30$ | $\mathbf{11 \times 20}$ | (66) | $\mathbf{11 \times 20}$ |
| Rel. pose $f$+E+$f$ 6pt | Kukelova *et al*. [25] | $31 \times 46$ | $31 \times 50$ | $31 \times 50$ | (218) | $\mathbf{21 \times 40}$ |
| Rel. pose E+$\lambda$ 6pt | Kuang *et al*. [24] | $48 \times 70$ | $34 \times 60$ | $34 \times 60$ | (846) | $\mathbf{14 \times 40}$ |
| Stitching $f\lambda$+R+$f\lambda$ 3pt | Naroditsky *et al*. [34] | $54 \times 77$ | $48 \times 66$ | $48 \times 66$ | (26) | $\mathbf{18 \times 36}$ |
| Abs. Pose P4Pfr | Bujnak *et al*. [7] | $136 \times 152$ | $140 \times 156$ | $\mathbf{54 \times 70}$ | (1745) | $\mathbf{54 \times 70}$ |
| Rel. pose $\lambda$+E+$\lambda$ 6pt | Kukelova *et al*. [25] | $238 \times 290$ | $149 \times 205$ | - | ? | $\mathbf{53 \times 115}$ |
| Rel. pose $\lambda_1$+F+$\lambda_2$ 9pt | Kukelova *et al*. [25] | $179 \times 203$ | $165 \times 200$ | $\mathbf{84 \times 117}$ | (6896) | $\mathbf{84 \times 117}$ |
| Rel. pose E+$f\lambda$ 7pt | Kuang *et al*. [24] | $200 \times 231$ | $181 \times 200$ | $\mathbf{69 \times 90}$ | (3190) | $\mathbf{69 \times 90}$ |
| Rel. pose E+$f\lambda$ 7pt (elim. $\lambda$) | - | - | $52 \times 71$ | $37 \times 56$ | (332) | $\mathbf{24 \times 43}$ |
| Rel. pose E+$f\lambda$ 7pt (elim. $f\lambda$) | Kukelova *et al*. [26] | $\mathbf{51 \times 70}$ | $\mathbf{51 \times 70}$ | $\mathbf{51 \times 70}$ | (3416) | $\mathbf{51 \times 70}$ |
| Abs. pose quivers | Kuang *et al*. [21] | $372 \times 386$ | $216 \times 258$ | - | ? | $\mathbf{81 \times 119}$ |
| Rel. pose E angle+4pt | Li *et al*. [31] | $270 \times 290$ | $266 \times 329$ | - | ? | $\mathbf{183 \times 249}$ |
| Abs. pose refractive P5P | Haner *et al*. [17] | $280 \times 399$ | $240 \times 324$ | $\mathbf{157 \times 246}$ | (8659) | $240 \times 324$ |

Table 1. Size of the elimination templates for some minimal problems. For the relative pose problems unknown radial distortion is denoted with $\lambda$ and unknown focal length with $f$, and the position describes which camera it refers to. The table shows the original template size from the author, the template size found using the method from [28] (GRevLex basis), the template size from doing an exhaustive search over Gröbner bases (Section 2.2) and the random sampling approach (Section 3.1). Missing entries are when the Gröbner fan computation took longer than 12 hours.

| Author | Execution time (ms) |
|---|---|
| Proposed | 0.16 |
| Larsson *et al*. [28] | 0.38 |
| Byröd *et al*. [8] | 0.89 |

Table 2. Timing of three point stitching with unknown focal length and radial distortion, using Matlab implementations running on a standard desktop computer.

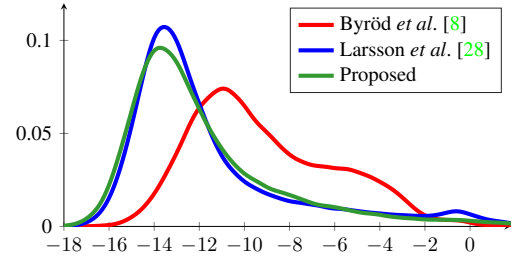

Figure 5. The figure shows histograms of equation residuals for 10,000 examples of the 3 pt stitching problem.

We have run both the exhaustive Gröbner basis selection and our proposed basis sampling scheme, see Table 1. The Gröbner bases do not give any improvement over the state-of-the-art solver but our sampling gives a significantly smaller template of size $18 \times 36$. Table 2 shows a comparison of runtimes.

### 4.2. Three View Image Stitching

The constraints (7) only compare pairs of images, using two point correspondences. So if we, instead of having three point correspondences in two views, have two point correspondences in three views, we get the same type of constraints, namely

$$\frac{\langle K^{-1}\boldsymbol{u}_1, K^{-1}\boldsymbol{u}_2\rangle^2}{|K^{-1}\boldsymbol{u}_1|^2|K^{-1}\boldsymbol{u}_2|^2} = \frac{\langle K^{-1}\boldsymbol{u}'_1, K^{-1}\boldsymbol{u}'_2\rangle^2}{|K^{-1}\boldsymbol{u}'_1|^2|K^{-1}\boldsymbol{u}'_2|^2}, \quad (8)$$

and

$$\frac{\langle K^{-1}\boldsymbol{u}''_1, K^{-1}\boldsymbol{u}''_2\rangle^2}{|K^{-1}\boldsymbol{u}''_1|^2|K^{-1}\boldsymbol{u}''_2|^2} = \frac{\langle K^{-1}\boldsymbol{u}'_1, K^{-1}\boldsymbol{u}'_2\rangle^2}{|K^{-1}\boldsymbol{u}'_1|^2|K^{-1}\boldsymbol{u}'_2|^2}, \quad (9)$$

where double primes are used for image three. We can hence use the exact same solver to solve this case. In this case we have a true minimal case, since we only get two constraints on $f$ and $\lambda$.

### 4.3. Evaluation

We have implemented our solver in MATLAB, where all image coordinate input and manipulation were done using mex-compiled C++ routines. In order to have a fair comparison of our method with [28] and [8], we modified their code so that the corresponding image coordinate manipulations also were done using mex-compiled code. The timing comparison is shown in Table 2, and one can see a clear speed-up. The solvers were run on a standard desktop computer. In order to check the numerical stability of our solver, we generated synthetic data, and evaluated the equation residuals. The results can be seen in Figure 5. In order to see how well our method works in practice, we did an automatic panoramic stitching of two images, with a fish-eye lens and unknown focal length, shown to the left in Figure 6. We then ran our solver in a standard RANSAC framework, with tentative correspondences based on SURF features and descriptors. The results can be seen to the right in Figure 6. Here the panoramic image was done without any blending in order to show the correctness of the stitching. The transformation used was based on the best RANSAC solution from our solver based on only three point correspondences, without any further bundle adjustment.

## 5. Relative Pose $E + f\lambda$

As another example we consider the relative pose problem where the calibration and distortion parameter are known for only one of the two cameras. The goal is to find a fundamental matrix $F$ and distortion parameter $\lambda$ that satisfy the epipolar constraints

$$[\hat{x}_i,\ \hat{y}_i,\ 1]\, F\, [x_i,\ y_i,\ 1 + \lambda(x_i^2 + y_i^2)]^T = 0, \quad (10)$$

as well as a focal length $f$, that makes

$$E = F\mathrm{diag}(f, f, 1) \quad (11)$$

an essential matrix. The problem is minimal with seven point correspondences and has 19 solutions. The first solver was presented by Kuang *et al*. [24] and was recently improved by Kukelova *et al*. [26].

### 5.1. Formulation of Kuang et al.

Now we give a brief overview of the formulation used in Kuang *et al*. [24]. The scale of the fundamental matrix is fixed by setting $f_{33} = 1$, and the epipolar constraints yield seven equations in the monomials

$$\{\lambda f_{13}, \lambda f_{23}, \lambda, f_{11}, f_{12}, f_{13}, f_{21}, f_{22}, f_{23}, f_{31}, f_{32}, 1\}. \quad (12)$$

Using the first six equations, Kuang *et al*. linearly eliminate the first two columns of the fundamental matrix[1]

$$[f_{11}, f_{12}, f_{21}, f_{22}, f_{32}, f_{33}]^T = G\, [\lambda f_{13}, \lambda f_{23}, \lambda, f_{13}, f_{23}, 1]^T \quad (13)$$

where $G \in \mathbb{R}^{6\times 6}$. Finally, the last equation expresses the monomial $\lambda f_{13}$ as a quadratic function $h(\lambda, f_{13}, f_{22})$, which gives the additional equation

$$\lambda f_{13} - h(\lambda, f_{13}, f_{23}) = 0. \quad (14)$$

Parametrizing the inverse focal length $w$, the essential matrix is given by $E = F\mathrm{diag}(1, 1, w)$, and it must satisfy the equations

$$2EE^TE - \mathrm{tr}(EE^T)E = 0, \quad \det(F) = 0. \quad (15)$$

This gives 11 equations in unknowns $w, \lambda, f_{13}$ and $f_{23}$. Using these equations, Kuang *et al*. [24] constructed a polynomial solver with a template of size $200 \times 231$.

---
[1]Note that here we have the focal length and distortion on the right side of $F$, while it was on the left in [24].

Computing the Gröbner fan, we found that there are 3190 different reduced Gröbner bases for this problem. Constructing solvers for all of these bases, we found a solver of size $69 \times 90$. Applying the random approach in Section 3.1, we did not find any better solver. While this solver is significantly smaller than the original solver from Kuang *et al*. [24] ($200 \times 231$), it is still slightly larger than the state-of-the-art solver from Kukelova *et al*. [26] ($51 \times 70$).

### 5.2. Formulation of Kukelova et al.

In [26] the authors present another formulation for this problem based on computing elimination ideals to eliminate both the radial distortion and focal length. Since the radial distortion makes the epipolar constraints non-linear they first employ a lifting technique to remove the non-linearity. They introduce new variables $y_1, y_2$ and $y_3$ and construct an extended fundamental matrix as in [4],

$$\hat{F} = \begin{bmatrix} f_{11} & f_{12} & f_{13} & y_1 \\ f_{21} & f_{22} & f_{23} & y_2 \\ f_{31} & f_{32} & f_{33} & y_3 \end{bmatrix}, \quad (16)$$

together with the equations $y_i = \lambda f_{i3}$. Now the epipolar constraints are linear constraints on $\hat{F}$,

$$[\hat{x}_i,\ \hat{y}_i,\ 1]\, \hat{F}\, [x_i,\ y_i,\ 1,\ x_i^2 + y_i^2]^T = 0. \quad (17)$$

Using the (now) linear constraints on $\hat{F}$ they parametrize it using four unknowns. Finally using the elimination ideal trick they eliminate both the focal length and radial distortion parameter to get new polynomial constraints on the elements on $\hat{F}$. Using these new equations, they were able to construct a solver with a template size $51 \times 70$.

We computed the Gröbner fan for this parametrization and found that there are 3416 reduced Gröbner basis. Among these we found no solver better than the GRevLex solver built by Kukelova *et al*. We also performed the random sampling approach without finding any improvement. This matches our intuition that for equation systems where the unknowns are balanced in the monomials, GRevLex performs very well.

### 5.3. Our Approach

Empirically we have seen that our basis selection approach works best when the monomials appear in some unbalanced way in the equations. In the parametrization from Kukelova *et al*. [26], the only unknowns are the nullspace parameters from the linear equations.

To get more imbalanced equations, we propose another formulation which is a combination of the two previous approaches. In particular, we use the elimination ideal trick to eliminate the focal length, but keep the radial distortion parameter as an unknown. This avoids the extra unknowns introduced by the lifting in (16). Using similar lin-ear eliminations as Kuang *et al*. [24], the fundamental matrix is expressed in $\lambda$, $f_{13}$ and $f_{23}$. Then, instead of directly parametrizing the focal length and adding the essential matrix constraints (15), we add the eliminated constraints for one-sided focal length from [26] which only depend on the elements of the fundamental matrix. Together with the constraint from (14), we get five equations in only three unknowns. Computing the Gröbner fan, we find 332 different reduced Gröbner bases. The best solver was of size $37 \times 56$. Finally using the random sampling approach we find a solver with an elimination template of size $24 \times 43$.

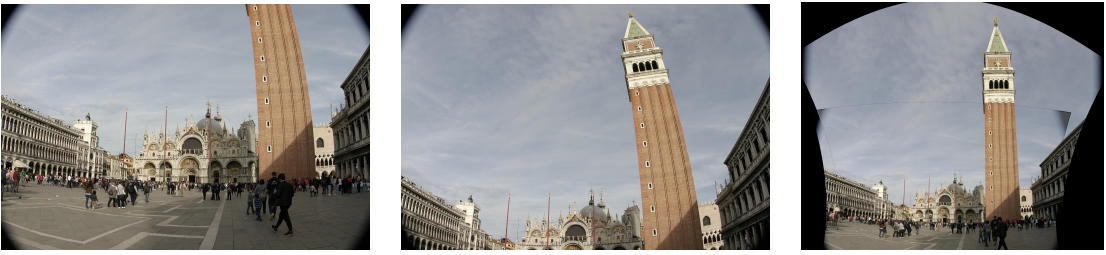

Figure 6. Stitching of two images with large radial distortion using on our three-point solver in a standard RANSAC framework. The resulting panorama (right) is based on the best RANSAC three-point solution without any additional non-linear refinement.

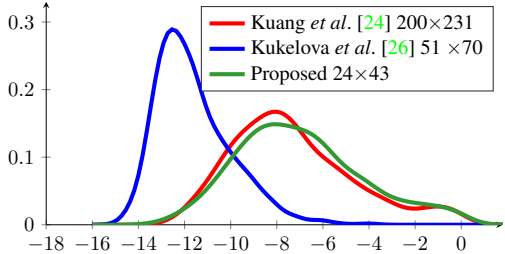

Figure 7. Relative error in focal length for 1,000 random instances.

### 5.4. Evaluation

We performed a synthetic experiment to evaluate the numerical stability of the new solver. We generated 1,000 random (but feasible) synthetic instances. The calibration parameters were set to $f_{gt} = 10$ and $\lambda_{gt} = -0.1$. For each solver we recorded the solution with the smallest focal length error. Figure 7 shows the distribution of the $\log_{10}$ relative focal length error $\frac{|f - f_{gt}|}{f_g t}$ for all 1000 instances. The numerical stability of the new solver is similar to the solver from Kuang *et al*. [24]. Note that while the stability is worse than the solver from Kukelova *et al*. [26], it is still stable enough for practical purposes. The new solver is however significantly faster with an average runtime of 1.2 ms, compared to 10 ms for the solver from Kukelova *et al*. [26] (both solvers are implemented in MATLAB). Note that this increase in speed is not only due to the smaller elimination template, but the coefficients in the template are less complex and cheaper to compute.

## 6. Conclusions

We have explored how basis selection can be used to make polynomial solvers based on the action matrix method faster. The concept of Gröbner fans is an efficient representation of the possible reduced Gröbner bases that arise from (infinitely many) different monomial orderings. This gives us a tool to enumerate and test all monomial bases that arise from different Gröbner bases. We have shown that this gives in some cases significantly smaller elimination templates, and hence much faster solvers. We have also introduced a novel sampling scheme, that optimizes some heuristic criteria that we have experimentally found to often give small templates. Our initial motivation for sampling was that the calculation and testing of all Gröbner fans in some cases takes very (or even unfeasibly) long time, but we found that going beyond Gröbner bases can yield even smaller templates. Our motivation has here been to optimize the template size but the framework could easily be modified to optimize other criteria. We have tested our method on a large number of minimal problems, and shown that we get significant speed-ups in many cases. We have also in more depth explored how our method can be used in two applications, namely panoramic stitching with unknown focal length and radial distortion and relative pose with unknown one-sided focal length and radial distortion.

## 7. Acknowledgements

This work is supported by the strategic research projects ELLIIT and eSSENCE, Swedish Foundation for Strategic Research project "Semantic Mapping and Visual Navigation for Smart Robots" (grant no. RIT15-0038), Wallenberg Artificial Intelligence, Autonomous Systems and Software Program (WASP) funded by Knut and Alice Wallenberg Foundation, the European Regional Development Fund under the project IMPACT No. CZ.02.1.01/0.0/0.0/15_003/0000468, and the Czech Science Foundation Project GACR P103/12/G084.